\def\year{2021}\relax
\documentclass[letterpaper]{article} 
\usepackage{aaai21}  
\usepackage{times}  
\usepackage{helvet} 
\usepackage{courier}  
\usepackage[hyphens]{url}  
\usepackage{graphicx} 
\usepackage{natbib}  
\usepackage{caption} 
\usepackage[dvipsnames]{xcolor}
\usepackage[utf8]{inputenc} 
\usepackage[T1]{fontenc}    
\usepackage{booktabs}       
\usepackage{amsfonts}       
\usepackage{nicefrac}       
\usepackage{microtype}      
\usepackage{amsmath,amssymb}
\usepackage{subcaption}
\usepackage{tikz}
\usepackage{pgfplots}
\usepackage{enumitem}
\pgfplotsset{compat=1.16}
\usepackage{xspace}
\usepackage{breqn}
\usepackage{capt-of}
\usepackage{booktabs}
\usepackage{varwidth}
\newcommand*{\transpose}{^\intercal}
\newcommand{\myparagraph}[1]{\noindent\textbf{#1}}
\def\hfillx {\hspace*{-\textwidth} \hfill}
\newcommand{\etal}{et~al.\@\xspace}
\newcommand*{\eg}{e.g.\@\xspace}
\newcommand*{\ie}{i.e.\@\xspace}
\newcommand*{\etc}{\@ifnextchar{.}{etc.}{etc.\@\xspace}}
\newcommand{\arglocmin}[3]{\mathrm{arg\;loc\;min}_{{#1},{#2}} {#3}}

\DeclareMathOperator*{\argmin}{arg\;min}

\begin{document}
\title{Identifying Wrongly Predicted Samples: A Method for Active Learning}
\author{Rahaf Aljundi  \, Nikolay Chumerin \,  Daniel Olmeda Reino  \\ \,\\Toyota Motor Europe}
%


%
%

\maketitle
\begin{abstract}
State-of-the-art machine learning models require access to significant amount of annotated data in order to achieve the desired level of performance. While  unlabelled data can be largely available and even abundant,   annotation process can be quite expensive and limiting. Under the assumption that some samples are more important for a given task than others,
active learning targets the problem of identifying the most informative samples that one should acquire annotations for.
Instead of the conventional reliance on model uncertainty as a proxy to leverage new unknown labels, in this work we propose a simple sample selection criterion that moves beyond uncertainty. By first accepting the model prediction and then judging its effect on the generalization error, we can better identify wrongly predicted samples. We further present an approximation to our  criterion that is very efficient and provides a similarity based interpretation. 
In addition to evaluating our method on the standard benchmarks of active learning, we consider the challenging yet realistic scenario of imbalanced data where categories are not equally represented. We show state-of-the-art results and  better rates at identifying wrongly predicted samples.
Our method is simple, model agnostic and relies on the current model status without the need for re-training from scratch.
\end{abstract}
\section{Introduction}\label{sec:introduction}
The success of deep learning relies on the availability of large annotated data in which models trained on millions of samples can exceed human level performance in certain tasks~\cite{AI2007Progress}, however, obtaining annotations for new data can be a time consuming and a very expensive procedure. Besides, in many applications, \eg, semantic segmentation, samples are not equally important for the task being learned. Many samples can be redundant or easily predicted 
and annotating  them can be a waste of resources. With the goal of improving the data labelling efficiency, active learning is a sub-field of machine learning that aims at identifying the most informative data points in {a stream or } a large  pool of unannotated samples. Those identified samples are then sent for annotation and added to the existing training data, which would contribute to a substantial performance gain upon retraining, a process that can be  repeated until reaching a certain level of performance or consuming the annotation budget.
While a large body of research has been dedicated to active learning (see~\cite{settles2009active} for a survey), fewer research works have considered active learning with deep models. 
Neural network based methods for selecting labelling candidates   can either rely on the uncertainty in their current predictions~\cite{gal2016dropout,wang2016cost,gal2017deep} or on how representative the selected candidates are to the rest of the  samples~\cite{sener2017active}, or, alternatively, how different selected samples  are from the current training data~\cite{sinha2019variational}.
The last two methods~\cite{sinha2019variational,sener2017active} show state-of-the-art results when extracting large batches of  samples  from balanced pools of unannotated data. However,  the first requires solving a mixed integer optimization problem over all  pool samples and the later trains a VAE and a discriminator on the pool and training samples prior to  the selection process.
In this work, we are interested in developing a lightweight sample selection method
, this has its advantage in scenarios with large  abundant unlabelled data.

Given this state of mind, the most attractive direction, on one hand, seems to be the uncertainty based approach. On the other hand, this approach requires to train a Bayesian neural network or at least training a network with dropout which might not always be applicable or favorable. In this work, we propose a model agnostic and simple  method to select most informative samples. 
We suggest that  informative samples are those that the current trained model has predicted their output (label) wrongly and by acquiring their true labels, the model will gain access to new  bits of knowledge. 
Now the question we aim at answering is if there is a better way of pointing at wrongly predicted samples other than the uncertainty in their predictions. We propose to identify  those wrongly predicted samples by first hypothesizing that the model prediction is correct and attempt at increasing the confidence of the model in its initial prediction. We then measure the effect of this hypothesis on the model performance on a small holdout set. As increasing the confidence on wrong predictions would harm the model performance in contrary to correct predictions, we use the relative change in the model performance (or alternatively the error) as a criterion for selecting samples that are likely to be wrongly predicted. 
Our method is generic, efficient and requires no changes on the current model.
\\
Aside from these aspects, in this work, we point at the fact that active learning methods have been mostly tested in settings where the pool of unannotated samples is artificially balanced over the different categories as in the case of most standard datasets.
 This assumption is unrealistic in many cases and hides the potential of the different approaches, \eg, random sampling is only outperformed by a small margin.
We argue that real life applications often face the problem of imbalanced set of samples and the condition where samples are balanced among different classes is solely met in existing benchmarks. In this paper, we consider the challenging setting of imbalanced pool of unannotated samples where not all categories are equally represented. We show that random selection is no longer a competitive baseline and requires significant extra amount of annotations in comparison to our method which targets regions where most mistakes occur and surpasses the imbalanced nature of the data.   While our proposed method has a comparable or better performance to its counterparts on the controlled balanced setting, we show a significant improvement on the more realistic scenario of imbalanced pool samples.\\
Our contributions are as follows: 1)~we propose a novel approach for sample selection based on their plausibility of being wrongly predicted by the current trained model. 2)~We present an approximated variant of our method and demonstrate an interesting link with kernel based similarity measures, here  from a network perspective. 
3)~We  achieve state-of-the-art results especially on the realistic yet challenging imbalanced setting. 
In the following, we  discuss closely  related works in Section~\ref{sec:related_work} and  describe our proposed approach in Section~\ref{sec:method}. We evaluate our method  on image classification problem, Section~\ref{sec:image_classification} and semantic segmentation problem, Section~\ref{sec:image_segmentation}, we conclude in Section~\ref{sec:conclusion}.
\section{Related Work}\label{sec:related_work}
{ Active learning is an important field of machine learning and has been studied extensively well before the success of deep learning, we refer to~\cite{settles2009active,fu2013survey} for surveys. Our work considers  pool based setting where  annotation candidates are to be selected from a big pool of unlabelled data. Under this setting,  most studied lines of work focus either on identifying current uncertain samples or alternatively a set of diverse and representative samples~\cite{fu2013survey}. However, our work comes closest to  approaches that aim at selecting samples which would, once annotated, incur the largest effect on the trained model. 
We mention the largest expected model change approach as in 
 EGL~\cite{settles2008multiple} where samples are selected based on an approximation to the expected value of the sample gradient given the current predicted output distribution.
Samples with largest gradient magnitude are selected for annotation. In  our method approximation, we don't depend  only on the gradients magnitudes, but also on the angle between the gradient of the pool sample  and estimated gradient of the holdout set.\\
 Instead of estimating the model change by the expect gradient length, variance reduction methods~\cite{hassanzadeh2011variance,hoi2006large} aim at implicitly reducing the generalization error by selecting candidates that would minimize the model output variance through the reliance on  Fisher information.
Closer to our approach, expected error reduction methods, estimate explicitly how much the generalization error is going to be reduced as in~\cite{roy2001toward}. 
For each candidate, the model is trained on each possible label and the  generalization error is computed on other pool samples, approximated with the current model output distribution and further averaged over the different possible labels of the candidate.  In our work, we also aim at reducing the generalization error of the model when the selected samples are correctly annotated, however, we select samples that affect most negatively the generalization error when using their current predicted labels as groudtruth. We use this as a proxy to identify wrongly predicted labels. 
 Aside from the novel deployed criterion, in this work, we introduce  a series of steps to make such approach applicable to deep learning.  Instead of estimating an expectation of the loss on the pool set, we deploy the typically required validation set to estimate the generalization error and instead of estimating the excepted updated model given all possible labels we rely on pseudo labels. 
 More importantly we present an efficient approximation and show how our criterion can be interpreted as selecting samples that are dissimilar to those in the holdout set.\\
}
When considering active learning methods designed for deep learning,  several paradigms have emerged such as uncertainty based sampling~\cite{gal2016dropout, gal2017deep,NIPS2017_7141}, representation based sampling~\cite{sener2017active,sinha2019variational} and query by committee using ensemble of models~\cite{gilad2006query,seung1992query}.
Uncertainty based methods are the most similar in nature to our approach. Gal~\etal in~\cite{gal2016dropout} showed that MC-Dropout can be used to perform approximate Bayesian inference in deep neural networks, and applied it to high-dimensional image data in~\cite{NIPS2017_7141}  to estimate uncertainty as a selection criterion. Our approach moves beyond uncertainty by selecting samples that are likely to be wrongly predicted using the change in generalization error. In~\cite{wang2016cost} the obtained annotations of uncertain samples and the pseudo labels of the most certain samples were combined. In our work, we provide a ranking of the samples, where the points that reduce generalization error can be deployed along with their pseudo labels for further performance improvement.
{ Very recently, \cite{ash2019deep} proposes to rely on  the gradient magnitude as a measure of uncertainty while selecting diverse candidates. In our approximation, we select samples that are dissimilar  to a holdout set in the gradient space.}
\section{Our Approach}\label{sec:method}
\begin{figure*}[t]
    \centering
    \subfloat[]{{\includegraphics[width=.24\textwidth]{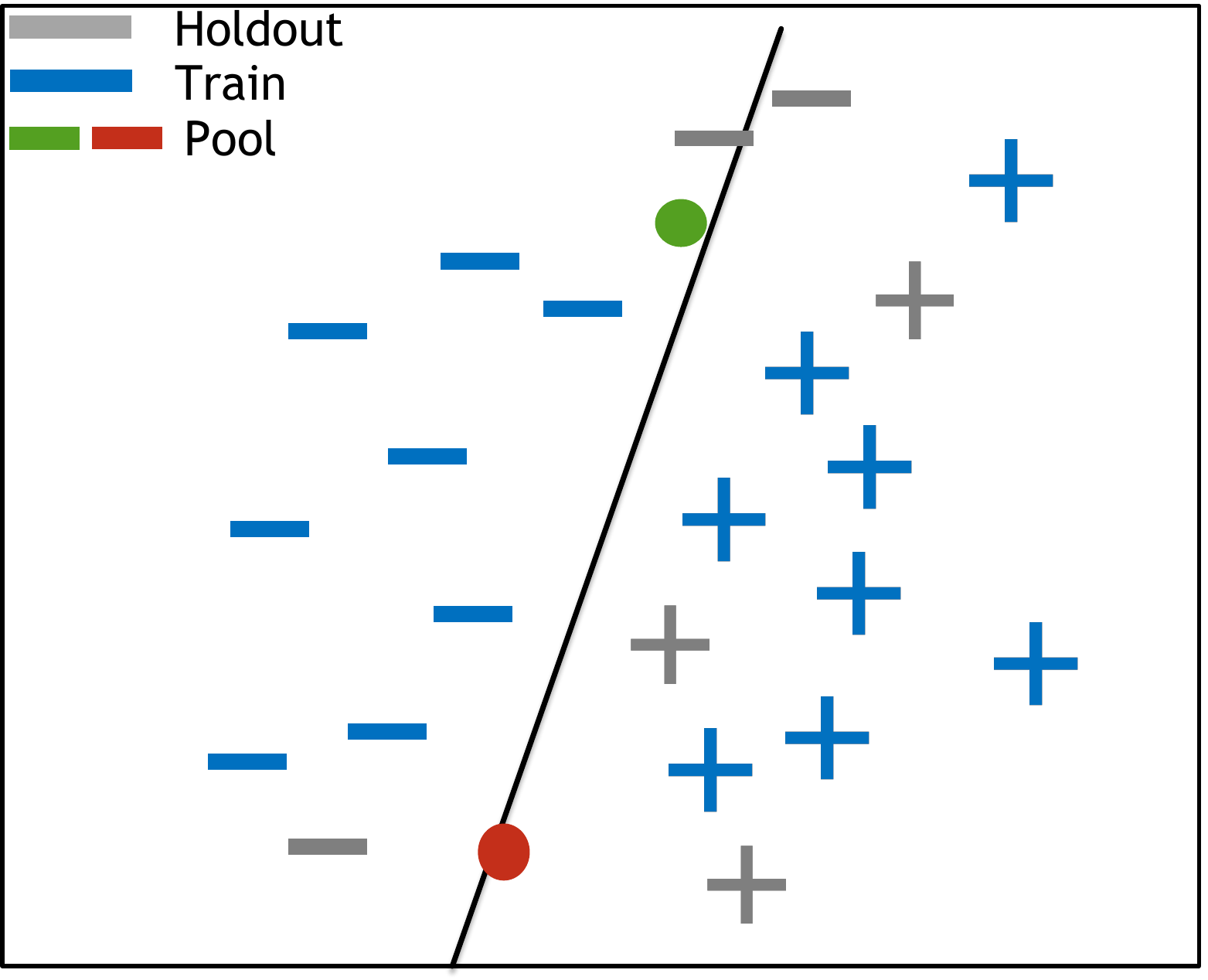}} }%
     \hfill
    \subfloat[]{{\includegraphics[width=.24\textwidth]{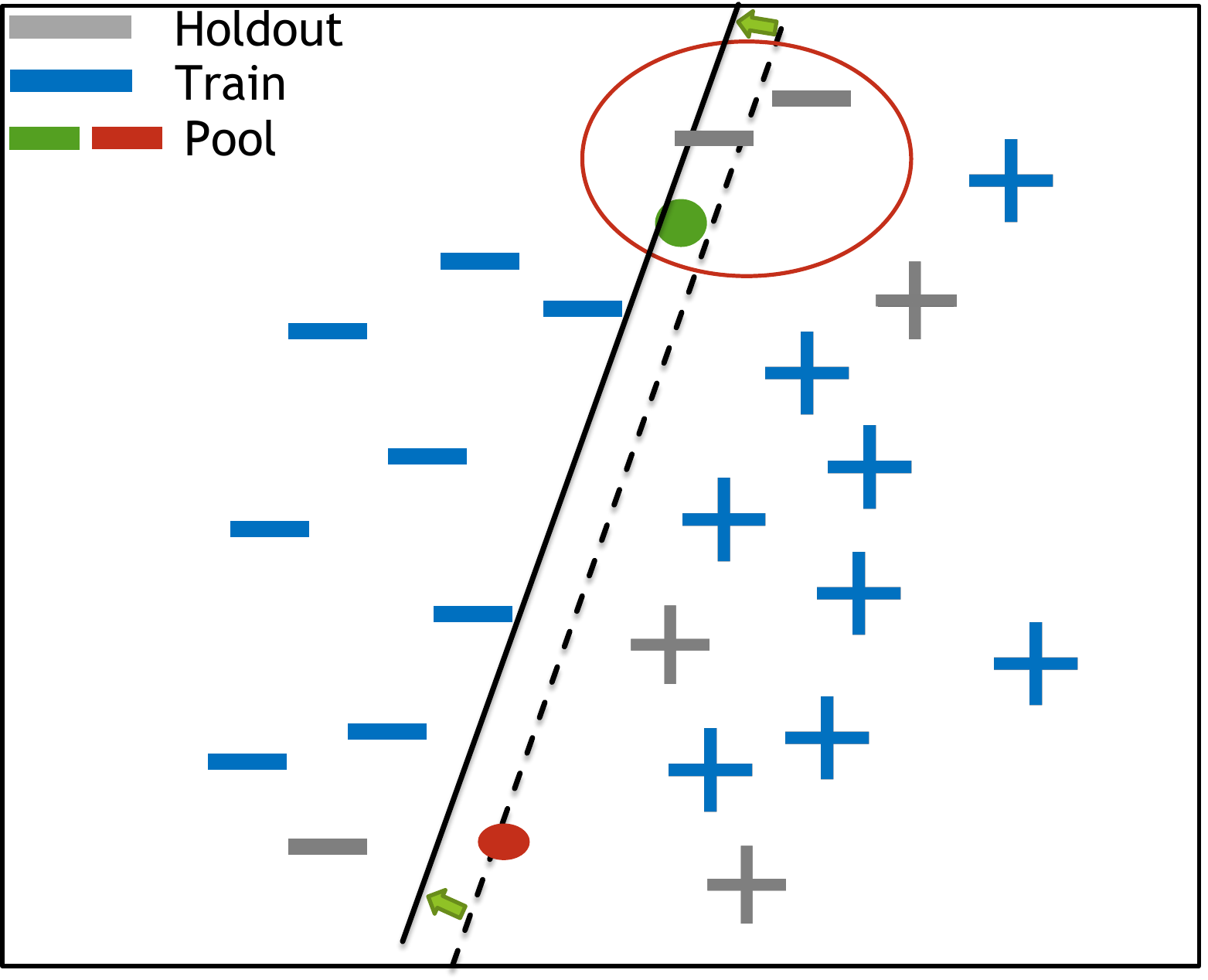} }}
        \hfill
    \subfloat[]{{\includegraphics[width=.24\textwidth]{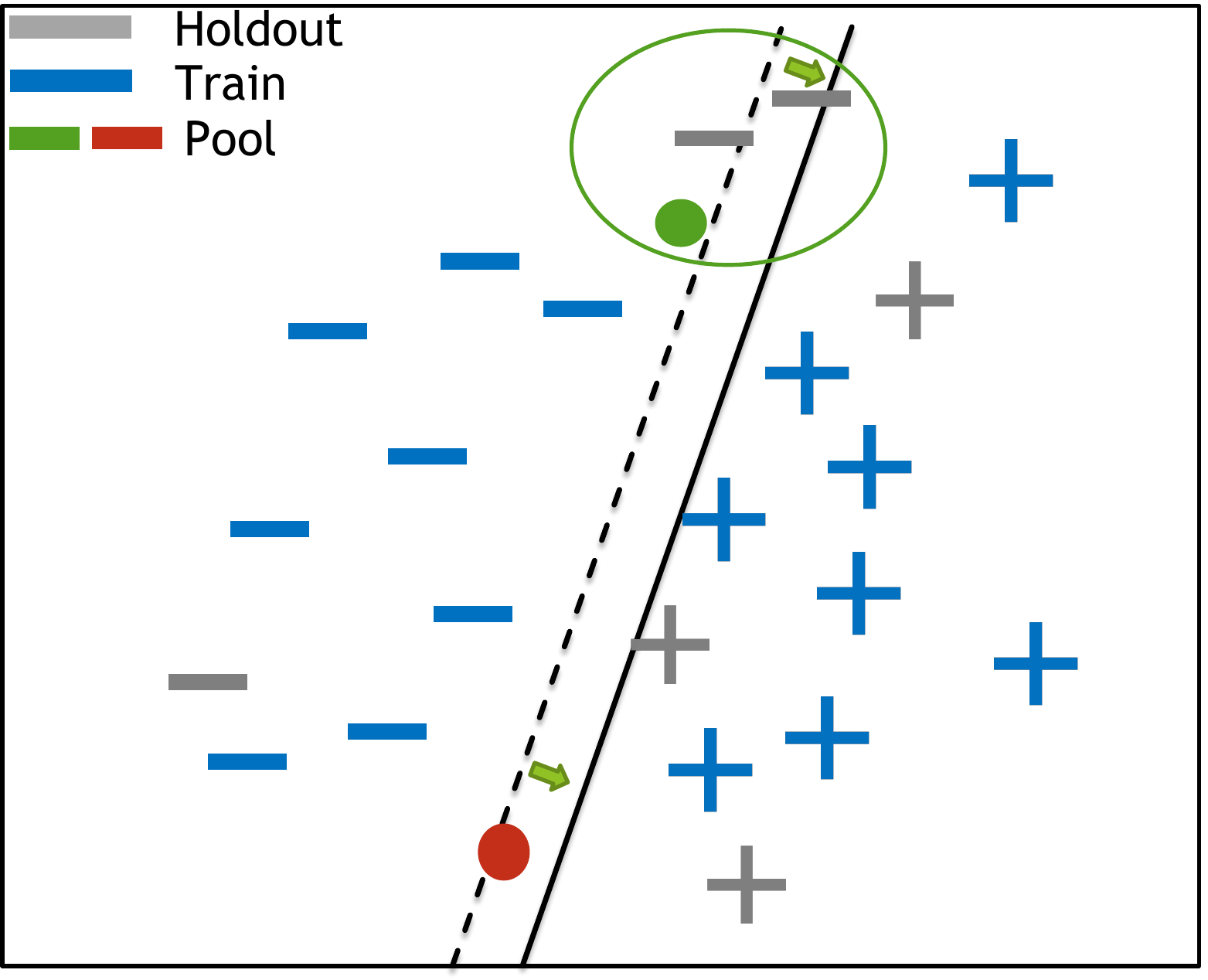} }}
       \hfill
    \subfloat[]{{\includegraphics[width=.24\textwidth]{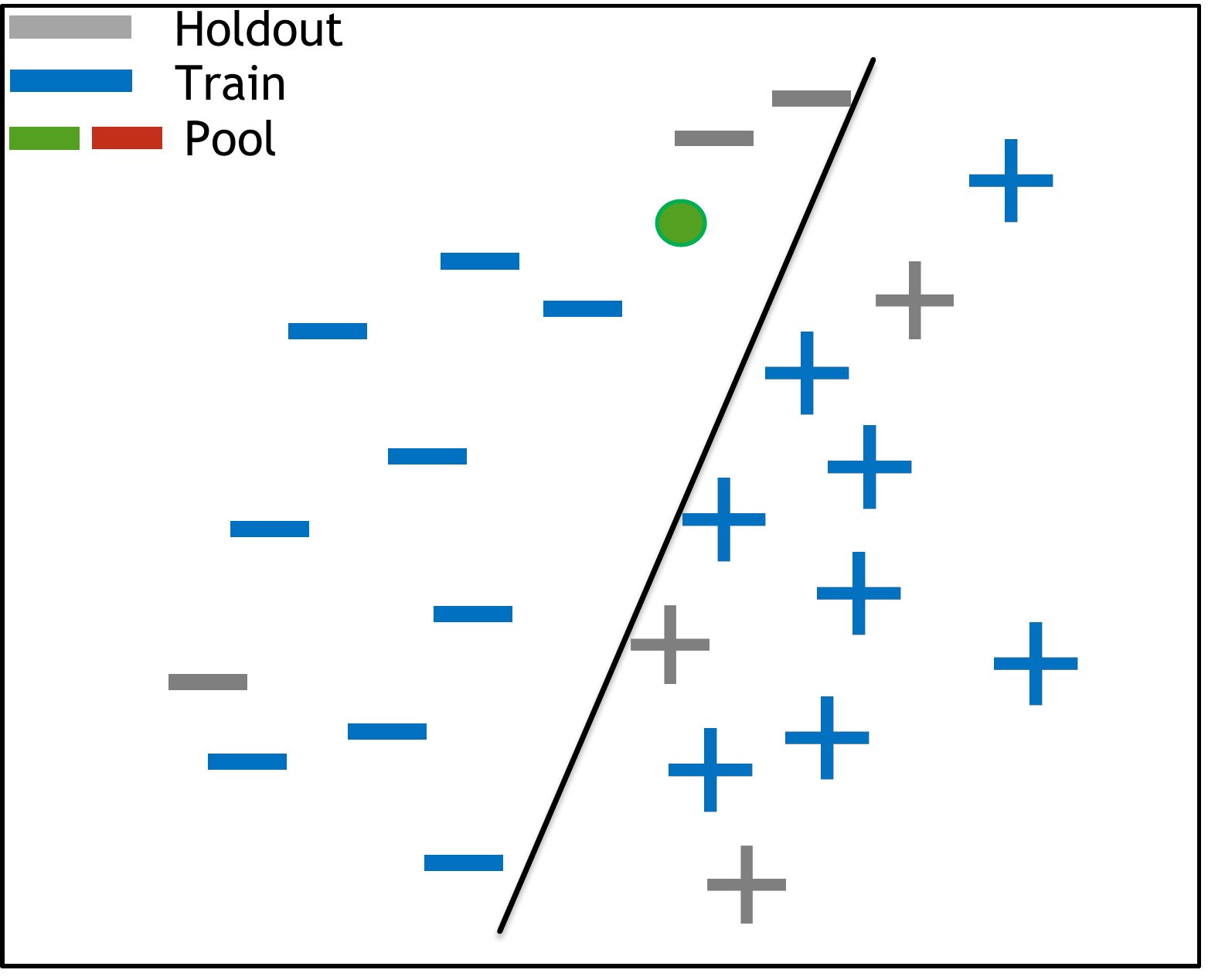} }}
\caption{\footnotesize Illustration of the main idea behind our approach (best viewed in color). (a)~Shows the initial training set in blue, the validation set in gray, used here as a holdout set, and the pool samples (circles) in addition to the current learned decision boundary. The two pool samples (with ground truth of negative labels) are uncertain with one (in green)  correctly predicted  and the other (in red)  wrongly predicted. (b)~Illustrates how the decision boundary would move if we make  small training step(s) with the prediction of the red pool sample as a true label,  while (c)~illustrates the effect of utilizing the correctly predicted output of the green pool sample instead. Using the (wrongly predicted) label of the red pool sample would harm the generalization performance as some correctly predicted samples will be misclassified as opposed to minimizing the loss on the correctly predicted sample. Hence, our method  selects the red point for annotation in order to obtain its correct label. Finally, (d)~shows the decision boundary after adding the newly annotated sample to the training set.}%
    \label{fig:illustration}%
    \vspace*{-0.2cm}
\end{figure*}
While standard supervised machine learning methods have access to all available training data for a given task before the start of the learning process, active learning methods assume access to only small initial set ${X^t}$ of labelled training data along with a much larger set of unannotated data ${X^p}$~(pool). The active learning method $\mathcal{M}$ should identify a set of $K$ most informative samples $X^a$ to be annotated and added to the training set which should contribute to a maximal gain in the trained model performance. $K$ is the size of the annotation step $s$.  This process can be repeated until reaching a certain performance level or exhausting annotation resources with $S$ annotation steps performed.  
Given a neural network parameterized by $\theta$, we want to learn a function $f(x;\theta)$ that maps the input data $x$ to their corresponding labels ${y}$. 
We start by training the model on the  initial training set $X^t$, considered as a starting step $s=0$:
\begin{equation}
 \theta^{s}=\argmin_\theta\,\ell\left(f(X^t;\theta),Y^t\right).
 \vspace{-0.15cm}
\end{equation}
The goal is to estimate each pool sample's utility, which is related to the amount of information conveyed while pairing the sample with the correct output.
Providing already known labels might not be  beneficial compared to  correcting  current mistakes. While a popular line of works~\cite{gal2016dropout,wang2016cost,gal2017deep} rely  on uncertainty, we argue that a model can be uncertain about a sample and yet predicts its output correctly. In this work, we propose a new method to spot wrong predictions.

We consider a standard classification problem, where the target class $y$ is estimated from the model output $f(x;\theta)$. 
Starting from the initial predictions of the model, the entropy of these  predictions 
can be used as a measure of uncertainty. Minimizing this entropy, which is usually deployed in unsupervised or semi-supervised learning~\cite{grandvalet2005semi,berthelot2019mixmatch,guerrero2000entropy}, would push the prediction towards the most probable label and suppress other labels probabilities. 
Assume that the predicted label is correct, then maximizing the confidence in this prediction through maximizing the log-likelihood of the predicted label or alternatively minimizing the cross-entropy loss using the first predicted label as a pseudo label could help improving  the model performance. However, if the initial model prediction is wrong then minimizing the  loss on a wrong label could harm the model performance as we are injecting false information in the model. We use this reasoning as a base for designing a measure to select wrongly predicted samples.
Figure~\ref{fig:illustration} illustrates the main idea behind our approach.\\
Formally, for each candidate sample $x_i^p\in X^p$,
we first obtain
its prediction $\hat{y}_i$ 
by the current model (with $\theta=\theta^s$). 
In unsupervised learning methods,
{$\hat{y}_i$} can be seen as a pseudo label and used to train the model~\cite{lee2013pseudo}. Here, we first assume that the prediction given by the pseudo label {$\hat{y}_i$} is correct and through minimizing the loss {$\ell$} on the current sample {$x_i^p$} given the pseudo label {$\hat{y}_i$} we obtain {$\theta^p_i$} as:
\begin{equation}\label{eq:min_loss_pool}
    \theta^p_i =
    \arglocmin{\theta}{\theta^s}{\ \ell\left(f(x_i^p;\theta),\hat{y}_i\right)}. \end{equation}
Here $\arglocmin{\theta}{\theta_s}{}$ denotes the argument-result of the local minimization w.r.t.~$\theta$  starting from~$\theta^s$. 
Now, as explained earlier, our hypothesis suggests that if the current model prediction is wrong, the minimization of the model loss given that prediction as a target label would harm the model performance.
This can be due to moving the decision boundary in the wrong direction or relying on a feature that is apparent or representative of another category. 
We use this rule as a proxy to identify the samples that are likely to be wrongly predicted.
We propose to select samples using the estimate of the change in the  model generalization error, \ie, the change in the model prediction error on unseen samples:
\begin{equation}\label{eq:pseudo}
\mathcal{S}(x_i^p) =\,
\ell\left(f(X^v;\theta^p_i),Y^v\right) - \ell\left(f(X^v;\theta^{s}),Y^v\right).
\end{equation}

To measure the generalization error, we employ a small set $X^v$, this can be a small holdout set or the validation set  used  for setting hyper-parameters and estimating the model performance. 
We then select {$K$} samples with largest values of $\mathcal{S}$ and request their annotations. The newly labelled data are to be added to the training set on which the model will be trained again and then a new active learning step can be carried out. 
It can  be noted that in general the last {$K$} samples whose updated model loss decreases are  likely to be correctly predicted by the current model and can be combined with the training pool to be learned in an "unsupervised" manner.\\
To summarize, instead of relying only on the model uncertainty to find the annotation candidates, we use the change in the generalization error on a holdout set to identify the wrongly predicted samples.  
\subsection{First Order Approximation}
Our  criterion involves estimating the updated model loss on a holdout set after the loss minimization on each pool sample given its  pseudo label. As we shall see in the experiments, Section \ref{sec:image_classification}, the holdout set can be very small and its loss can be estimated in one forward pass. Nonetheless, to account for scenarios with extreme constraints on computational cost and more importantly to gain  better insights on  our proposed method behaviour,  we analyze and present an approximation to the selection criterion in~\eqref{eq:pseudo}.
Let's define:
\begin{equation}
\ell_v(\theta) = \ell\left(f(X^v;\theta),Y^v\right)
=\sum_j \ell\left(f(x_j^v;\theta),y_j^v\right).
\end{equation}
Then~\eqref{eq:pseudo} can be rewritten as follows:
\begin{equation}
\mathcal{S}(x_i^p) = \ell_v(\theta^p_i) - \ell_v(\theta^{s}).
\end{equation}
We expand the first term about $\theta^s$ using first order Taylor series approximation. 
\begin{equation}\label{eq:approx_lv}
{
\ell_v(\theta^p_i) \approx \ell_v(\theta^s) + \nabla_\theta\ell_v(\theta^s)\cdot\left(\theta^p_i - \theta^s\right),
} 
\end{equation}
{
where $\nabla_\theta\ell_v(\theta^s)$ denotes the gradient of $\ell_v(\theta)$ w.r.t. the parameter $\theta$ at point $\theta=\theta^s$.
} 
Instead of doing  full optimization of the loss (mentioned   in~\eqref{eq:min_loss_pool}), we propose to estimate $\theta_i^p$ by a single step of gradient descent from $\theta^s$ with a learning rate $\eta$ and the loss gradient estimated at sample $x_i^p$:
\begin{equation}\label{eq:approx_theta_ip}
    \theta^p_i\approx\theta^s-\eta{\nabla_\theta\ell(f(x_i^p;\theta^s),\hat{y}_i)}.
\end{equation}
Then, using this estimate in the right-hand part of~\eqref{eq:approx_lv}, we obtain:
{\small
\begin{multline}
\ell_v(\theta^p_i)\approx
\ell_v(\theta^s) +{\nabla_\theta\ell_v(\theta^s)}\cdot\left(\theta^s-\eta{\nabla_\theta\ell(f(x_i^p;\theta^s),\hat{y}_i)}-\theta^s\right) \\
= \ell_v(\theta^s) -\eta
{\nabla_\theta\ell_v(\theta^s)}\cdot
{\nabla_\theta\ell(f(x_i^p;\theta^s),\hat{y}_i)},
\end{multline}
}
\vspace*{-0.6cm}
\begin{multline}\label{eq:Sxip-approximation}
\mathcal{S}(x_i^p) \approx \ell_v(\theta^s) 
-\eta\nabla_\theta\ell_v(\theta^s)\cdot\nabla_\theta\ell(f(x_i^p;\theta^s),\hat{y}_i) -\ell_v(\theta^s) \\
= -\eta\nabla_\theta\ell_v(\theta^s)\cdot\nabla_\theta\ell(f(x_i^p;\theta^s),\hat{y}_i).
\end{multline}
The positive constant $\eta$ in~\eqref{eq:Sxip-approximation} has no influence on the order of  $x_i^p$ sorted by decreasing $\mathcal{S}(x_i^p)$, and, therefore, can be simply dropped.
We propose the following alternative criterion and consider both criteria in the experiment Section~\ref{sec:image_classification}.
\begin{equation}\label{eq:approximated_criterion}
\mathcal{S}_a(x_i^p) = -{\nabla_\theta\ell_v(\theta^s)}\cdot{\nabla_\theta\ell(f(x_i^p;\theta^s),\hat{y}_i)}.
\end{equation}
The estimation of the selection criterion $\mathcal{S}_a(x_i^p)$ does not involve the loss minimization of~\eqref{eq:min_loss_pool} as in the original criterion $\mathcal{S}(x_i^p)$, but uses only the estimation of the pool sample gradient. It also replaces the estimation of the  holdout set loss in~\eqref{eq:pseudo} for each pool sample with a single prior computation of the loss gradient on the holdout set.
\subsubsection{A Similarity Based Interpretation.}
Here we want to present our criterion as a measure of  dissimilarity between a given pool sample and  samples of the holdout set based on the currently trained model.  Let's define the following kernel:
\begin{equation}\label{eq:our_kernel}
    K_\theta(x_i,x_j)=\nabla_\theta\ell(f(x_i;\theta),y_i)\cdot\nabla_\theta\ell(f(x_j;\theta),y_j).
\end{equation}
Each term in~\eqref{eq:our_kernel} can be expanded using chain rule as follows: 
\begin{equation}
  {\nabla_\theta\ell(\theta)}=\left(\nabla_\theta f(x_i;\theta)\right)\transpose\ell^\prime  ,
\end{equation}
where $\ell^\prime \in\mathbb{R}^C$ denotes the derivative of $\ell$ w.r.t. $f(x_i;\theta)$, and $\nabla_\theta f(x_i;\theta)\in\mathbb{R}^{C\times P}$ with $C$ the number of categories and $P$ the size of the parameter vector.
The kernel can be written as follows:
\begin{equation}\label{eq:our_kernel_expanded}
\resizebox{.88\hsize}{!}{$
    K_\theta(x_i,x_j)=\left({\left(\nabla_\theta f(x_i;\theta)\right)\transpose\ell^\prime}\right)\cdot\left({\left(\nabla_\theta f(x_j;\theta)\right)\transpose\ell^\prime}\right).$}
\end{equation}
This kernel is related to the Neural Tangent Kernel (NTK)~\cite{jacot2018neuraltangetkernel},  $K_{NTK}(x_i,x_j)=\nabla_\theta f(x_i;\theta)\cdot\nabla_\theta f(x_j;\theta)$, studied from an optimization point of view  in the infinite width limit, and describes how changing the network function at one point would affect its output on another.
A kernel similar to NTK was proposed in~\cite{charpiat2019inputsimilarity} to measure the similarity between samples from a trained network perspective for dataset self denoising. In the same study it was shown that from $\nabla_\theta f(x_i;\theta)=\nabla_\theta f(x_j;\theta)$ follows $f(x_i;\theta)=f(x_j;\theta)$, and samples with dissimilar features have orthogonal gradient directions and  kernels value close to zero. The kernel, defined in~\eqref{eq:our_kernel_expanded}, considers additionally the gradient of the loss function accounting for the sample/label pair. For example, in the case of a cross-entropy loss, we have  $K_\theta(x_i,x_j)=\left(\nabla_\theta f(x_i;\theta)\transpose\left(p_i -y_i\right) \right)\cdot  \left(\nabla_\theta f(x_j;\theta)\transpose(p_j -y_j)\right)$ with $y_i$ here constructed as a one-hot label vector and $p_i$ the output probability, which can be seen as weighting the function gradient by the difference between the predicted class probabilities and  the target/pseudo labels. See  supplementary materials for  details on the derivation.  
 Finally, given that 
$
\nabla_\theta\ell_v(\theta)=\sum_j \nabla_\theta\ell(f(x^v_j;\theta),y^v_j)$, 
 where $j$ is the index of the holdout samples,
our approximated criterion can be rewritten as follows:
\begin{equation}
\mathcal{S}_a(x_i^p) =-\sum_j K_\theta(x^p_i,x^v_j).
\end{equation}
Following that, our criterion allows to select the samples that are dissimilar to those in the holdout set according to the kernel defined in~\eqref{eq:our_kernel}.

\myparagraph{Binary classification example.}
Let us demonstrate the proposed sample selection method on a binary classification problem employing a single-layer neural network parametrized by ${\theta}$.
Assume the input to the network is a feature vector $\phi(x)\in\mathbb{R}_{\geq 0}^{n}$ extracted from a sample $x$ with a fixed (non-trainable) feature extractor $\phi$.
The function being learned is $f_\theta(\phi(x))=\theta\transpose\phi(x)$.
By defining $z=f_\theta(\phi(x))$, and the loss  $\ell(z,y)=-y\log(\sigma(z)) - (1-y)\log(1-\sigma(z))$, where $y\in\{0,1\}$ is the binary label, and $\sigma(z)=\frac{1}{1+e^{-z}}$;
the gradient of the loss w.r.t. $\theta$ can be derived using chain rule:
\begin{equation}
\nabla_\theta\ell(z,y) = \frac{\partial\ell(z,y)}{\partial\theta} =
\frac{\partial\ell}{\partial\sigma} \frac{\partial\sigma}{\partial z}
\frac{\partial z}{\partial\theta} =
(\sigma(z) -y)\phi(x).
\end{equation}
Following this definition, 
our criterion for selecting pool samples is:
{
\begin{equation}\resizebox{.88\hsize}{!}{$
{
    \mathcal{S}_a(x_i^p) =-\sum_j (\sigma(z^p_i) -\hat{y}_i)\phi(x_i)\cdot(\sigma(z^v_j) -y^v_j)\phi(x^v_j).
} $}
\end{equation}}
Let us analyze the kernel value w.r.t. a pool sample $x^p_i$ and a sample from the holdout set $x^v_j$. 
\begin{dmath}
    K_\theta(x^p_i,x^v_j)=c\, \phi(x^p_i)\cdot\phi(x^v_j),
    \text{with scalar }\; c=(\sigma(z^p_i) - \hat{y}_i)(\sigma(z^v_j) - y^v_j).
\end{dmath}
Consider the following cases: 
1)~the feature vectors $\phi(x^p_i)$ and $\phi(x^v_j)$ are different and $\phi(x^p_i)\cdot\phi(x^v_j)\approx0$, resulting in $K_\theta(x^p_i,x^v_j)\approx0$;
2)~$x^p_i$ and $x^v_j$ are close in the feature space and $\phi(x^p_i)\cdot\phi(x^v_j)\gg0$.
In the latter case, either $\hat{y}_i\neq y^v_j$ and $c<0$, causing $K_\theta(x^p_i,x^v_j)\ll0$ (case~2a), or $\hat{y}_i=y^v_j$ and $c>0$, leading to $K_\theta(x^p_i,x^v_j)\gg0$ (case~2b).
Therefore, if $x^p_i$ differs significantly from all holdout samples (case~1), then $\mathcal{S}_a(x_i^p)\approx 0$. However, if $x^p_i$ is similar to some holdout samples and it is predicted incorrectly (case~2b), then $\mathcal{S}_a(x_i^p)$ is likely to be positive, otherwise, when the prediction is correct (case~2b), the corresponding $\mathcal{S}_a(x_i^p)$ is likely to be negative.
Consequently, the pool samples from  both former cases would get greater (than the samples from the later case) values of the selection criterion and, therefore, will be selected for annotation.\\
In a nutshell, our method aims at selecting pool samples that differ from the holdout samples firstly in their (probably wrongly) predicted label or  in their feature representation.
\section{Experiments}
 To evaluate the effectiveness of our approach in various active learning scenarios, we perform a wide set of experiments on both image classification (Section~\ref{sec:image_classification}), and image segmentation   (Section~\ref{sec:image_segmentation}).

\subsection{Image Classification}\label{sec:image_classification}
\begin{figure*}[h!]
 \vspace{-.1cm}
    \centering
    \subfloat[ \label{fig:res_MNIST}]{{\includegraphics[width=.36\textwidth]{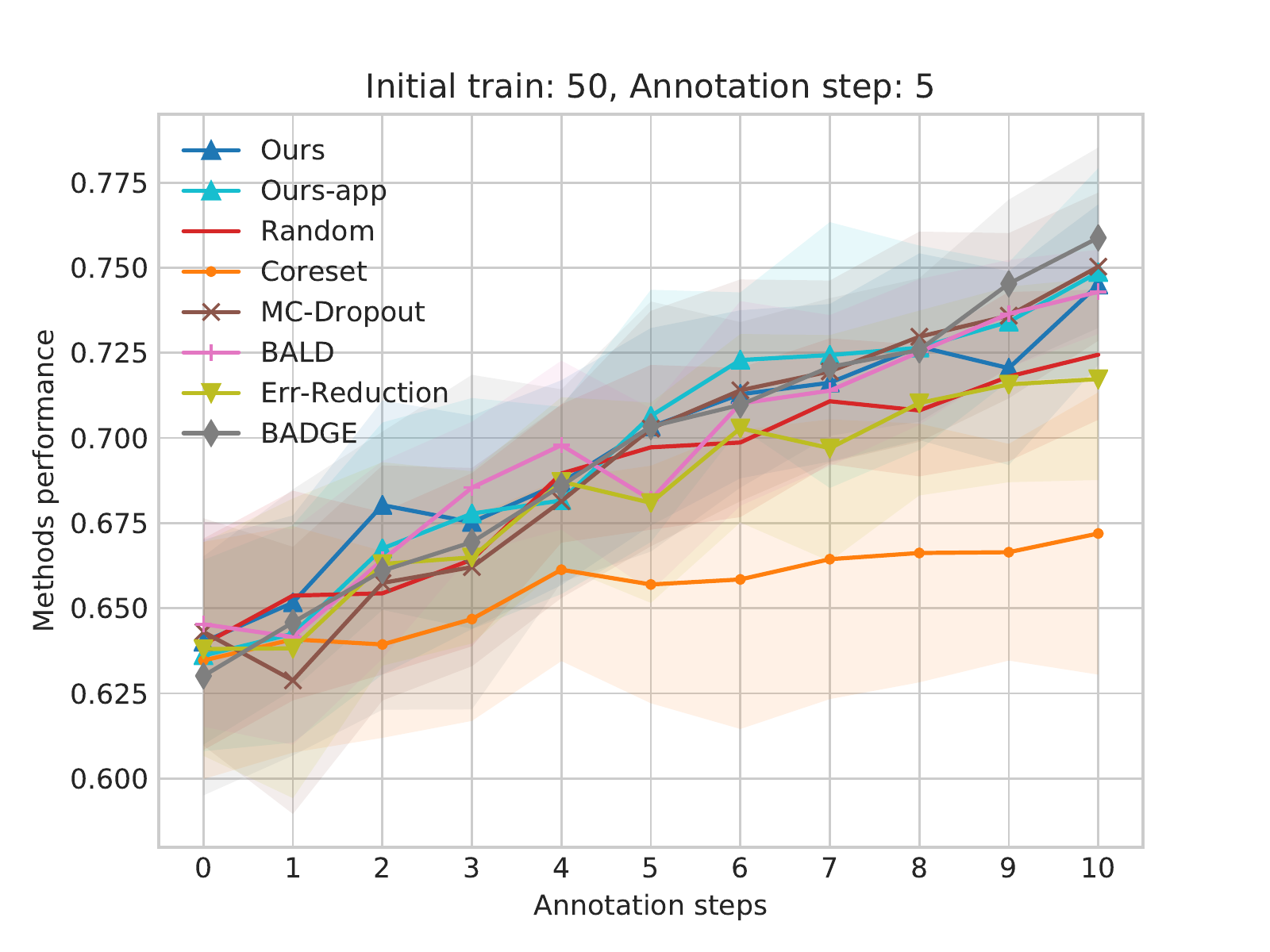} }}%
  \hfillx
    \subfloat[ \label{fig:res_KMNIST}]{{\includegraphics[width=.36\textwidth]{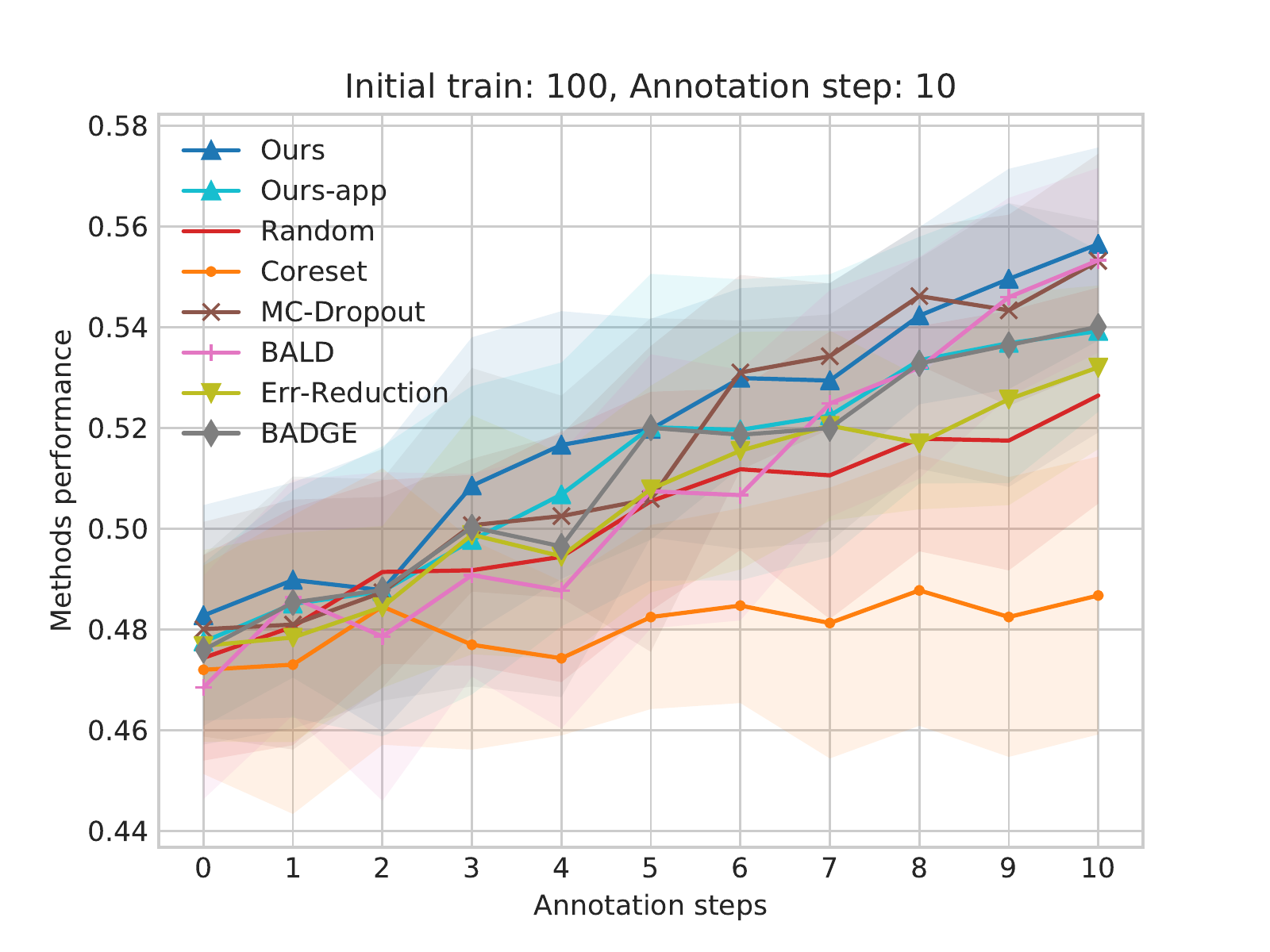} }}%
     \hfillx
        \subfloat[ \label{fig:res_SVHN}]{{\includegraphics[width=.36\textwidth]{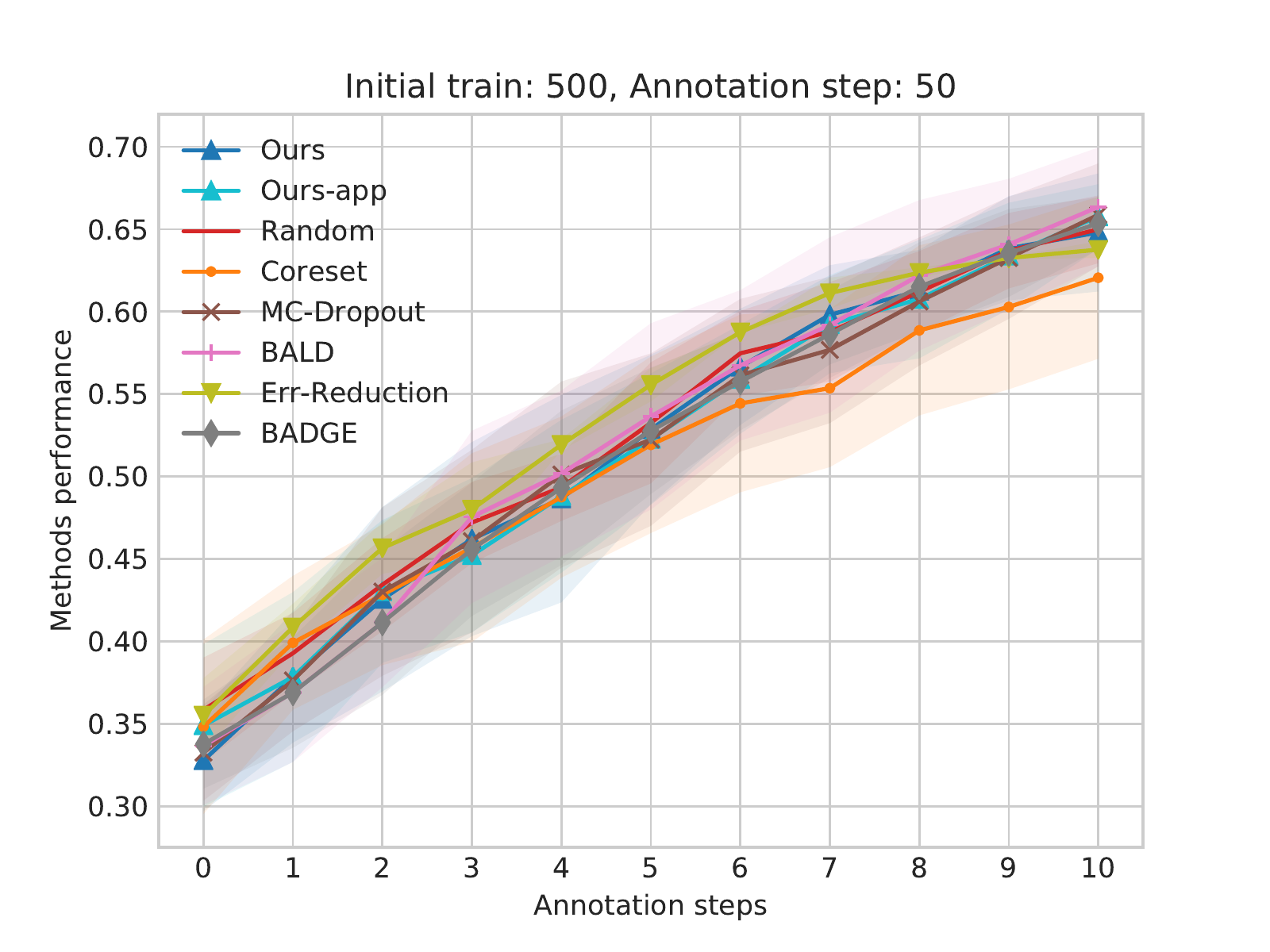} }}%
    \vspace{-0.2cm}
        \caption{\footnotesize Mean accuracy and std.dev. for (a)  MNIST, (b)  KMNIST and (c) SVHN on balanced setting.}%
     \vspace{-.1cm}
 \end{figure*}
We first compare our proposed method and its first order approximation  with state-of-the-art methods on the standard setting where datasets are balanced and all categories are represented equally.
We then move to a closer simulation of real life applications where we construct an active learning benchmark with imbalanced pool and initial training data.\\
\textbf{Datasets.}
On both settings we consider 
 MNIST~\cite{lecun1998gradient} dataset for handwritten digit recognition
, KMNIST~\cite{clanuwat2018deep} an MNIST style dataset for Kanji characters composed of 10 classes,
     SVHN~\cite{netzer2011reading} Google street view house numbers dataset and
     Cifar10~\cite{krizhevsky2009learning}.\\
\textbf{Compared methods.}
    - {\tt Random}: a random subset of the pool is selected for annotation at each step.\\
    { -{\tt Err-Reduction}~\cite{roy2001toward}}: an implementation of the error reduction approach using pseudo labels and error estimation on subset of the pool.\\
   - {\tt MC-Dropout}~\cite{gal2016dropout}: uses as a criterion the model uncertainty of each pool sample.\\ 
  -  {\tt Coreset}~\cite{sener2017active}:   selects a set of representative  samples covering the rest of the pool. 
  The method presents a mixed integer programming solution and a greedy alternative that is only $1-2\%$ inferior, we employ this efficient alternative.\\
  -  {\tt BALD}~\cite{gal2017deep}: it is  based on the mutual information between the  prediction  and the model posterior.\\
  {-{\tt BADGE}~\cite{ash2019deep} selects  diverse and uncertain samples in a gradient space based on the pseudo labels of pool samples. \\ 
  }
\textbf{Implementation.}
We deploy a two-layer fully connected network for MNIST and KMNIST datasets, and ResNet18 for SVHN and Cifar10.
All methods were trained using ADAM with early stopping on the  validation set. We don't retrain the model from scratch after each annotation step, we rather continue training  the model and only reset the optimizer parameters.  This makes more sense since the new samples are selected based on a criterion linked to the previously trained model. We apply this to all methods and observe consistent improvement. For our method, we consider both the criterion in~\eqref{eq:pseudo} and refer to it as {\tt Ours}, and the first order approximation criterion in~\eqref{eq:approximated_criterion} and refer to it as {\tt Ours-app}.
We  use the initial validation set  as a holdout set to  estimate the  criterion of both {\tt Ours} and {\tt Ours-app}.  Note that in our experiments we keep the validation set fixed to the initial setting while in practice one can augment it as new labels are obtained. We limit the optimization of  the loss in~\eqref{eq:min_loss_pool} to the last layer parameters and similarly for the gradients estimation of~\eqref{eq:approximated_criterion} of {\tt Ours-app} criterion. This has a valuable advantage computationally and shows no significant effect on the performance, see supplementary materials. The minimization of~\eqref{eq:min_loss_pool} in~{\tt Ours} is performed with SGD and limited to $3$ iterations with learning rate $\eta=10^{-3}$ on the fully connected model and $\eta=10^{-4}$ on ResNet. 
 We report the average of~$10$  runs with different random seeds  along with the standard deviation.
  \begin{figure*}[h]
  \vspace{-0.3cm}
    \centering
    \subfloat[ \label{fig:res_imb_MNIST}]{{\includegraphics[width=.36\textwidth]{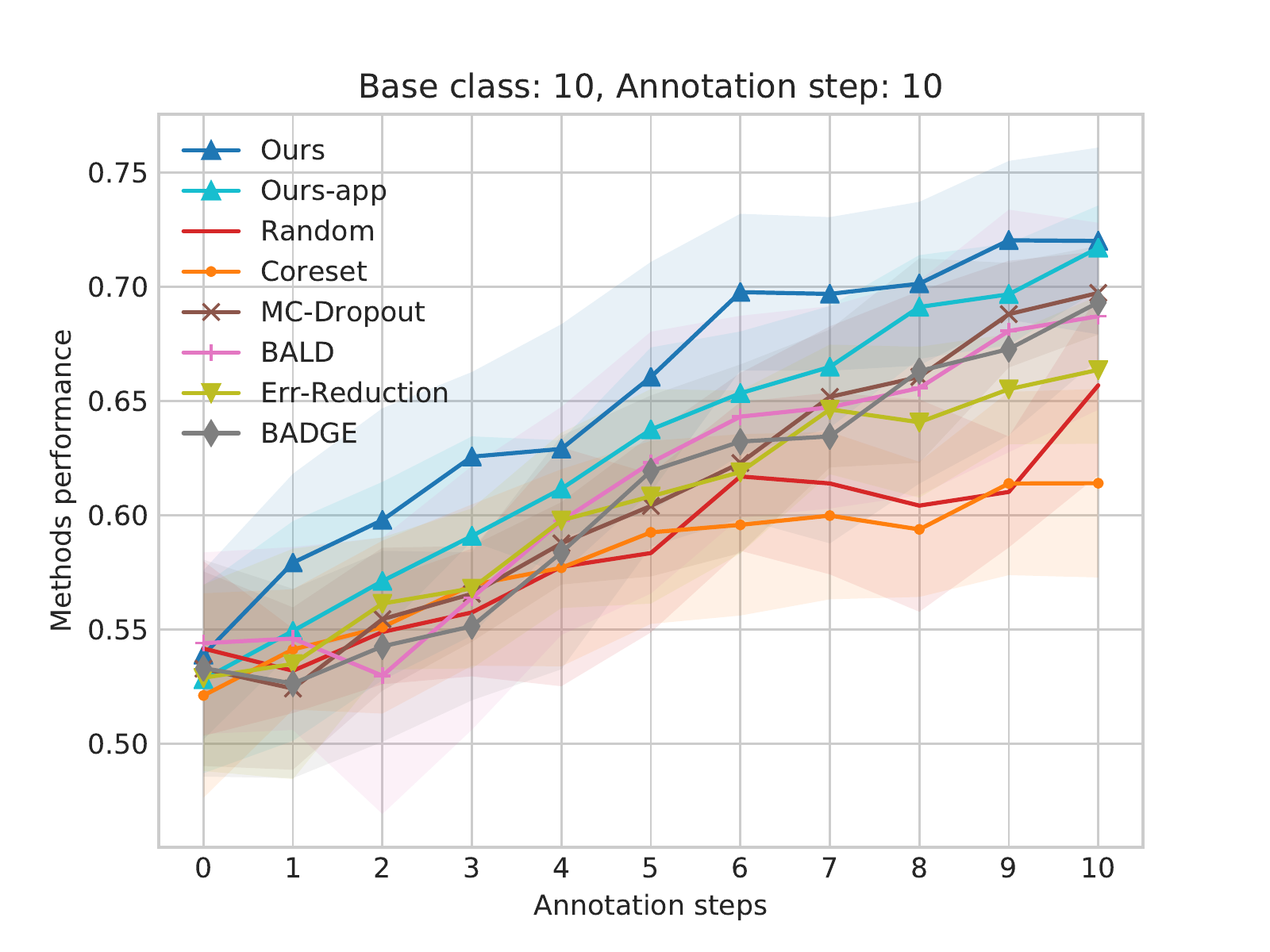} }}%
    \hfillx
    \subfloat[ \label{fig:res_imb_KMNIST}]{{\includegraphics[width=.36\textwidth]{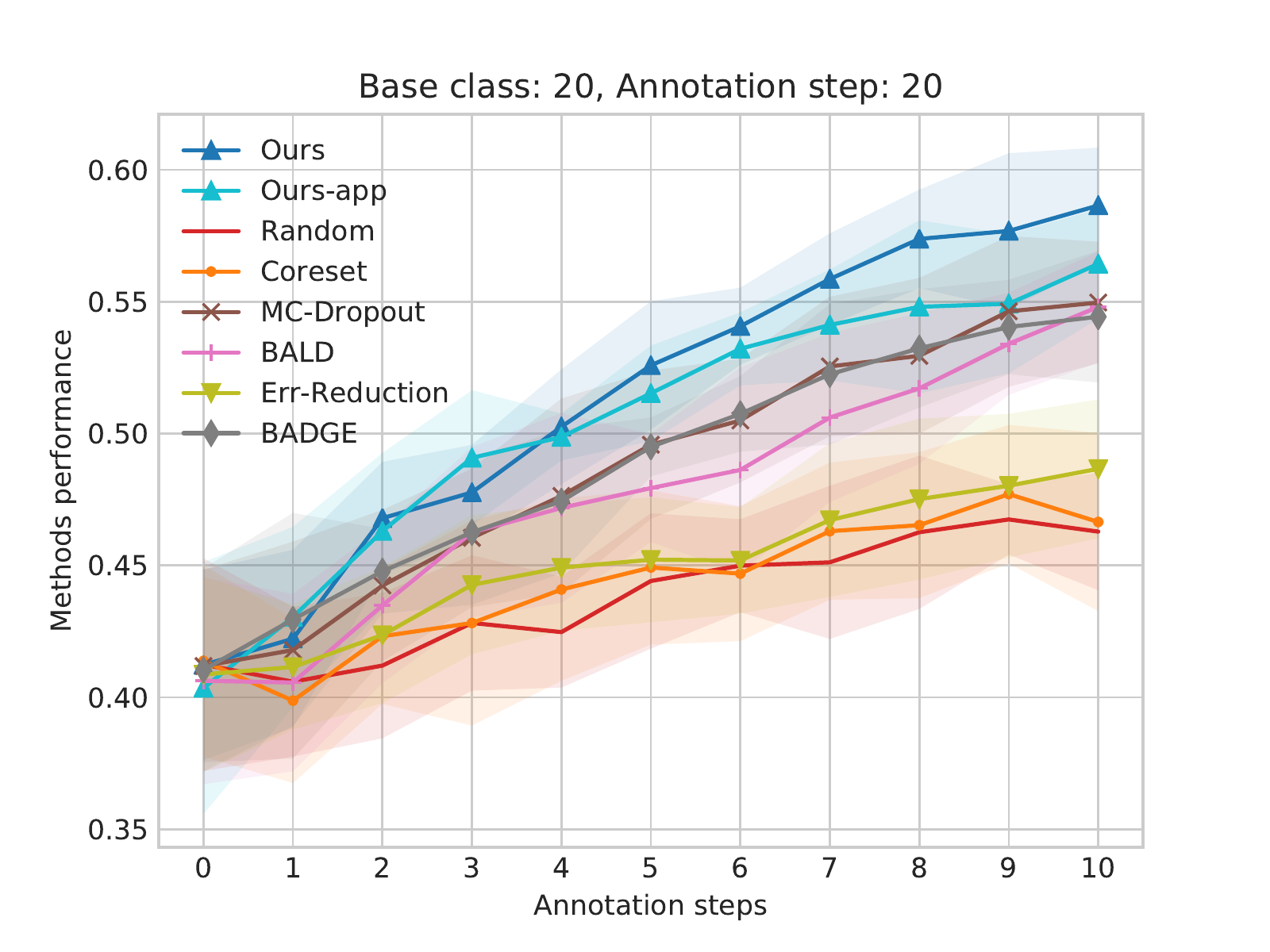} }}
        \hfillx
    \subfloat[ \label{fig:res_imb_SVHN}]{{\includegraphics[width=.36\textwidth]{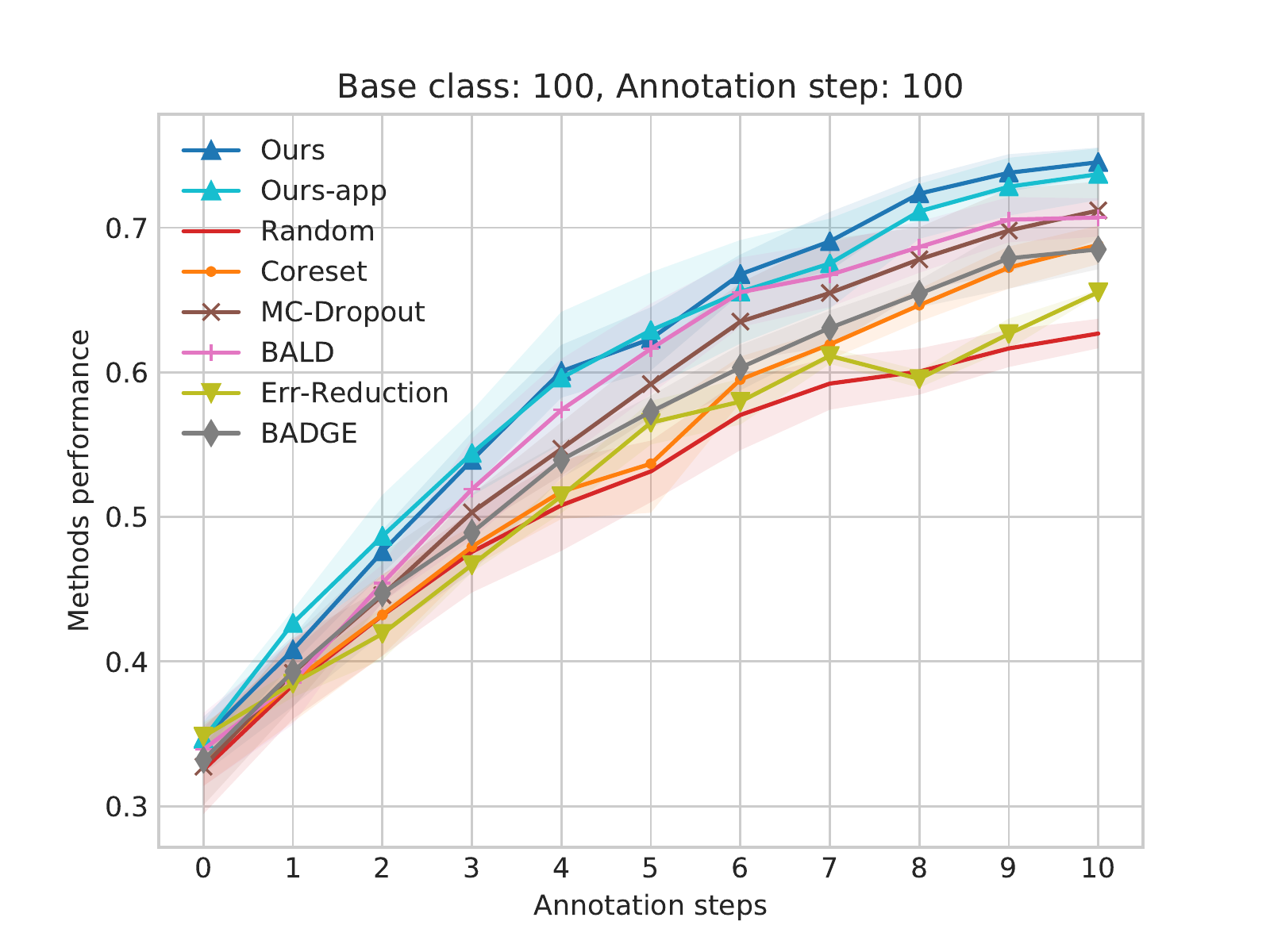} }}%
    \vspace*{-0.4cm}
   \caption{ \footnotesize Mean accuracy and std.dev.  for (a)  MNIST, (b)  KMNIST and (c) SVHN on imbalanced setting.}%
    \vspace{-0.4cm}
  \end{figure*}
\subsubsection{Balanced Setting.}
The used datasets are standard datasets in the field and they are composed of similar number of samples for each category. As such, the pool and the randomly sampled initial training set  represent equally each category. We refer to this setting as balanced setting.  Here, we use a validation set of the same size as the initial training set. Each active learning round is composed of 10 annotation steps with each step being $10\%$ of the initial training set size. The initial training set size  is relative to each dataset difficulty and the amount of samples needed to obtain a reasonable initial performance. We use the following initial  sizes 50, 100, 500, 5000 for MNIST, KMNIST, SVHN and Cifar10  respectively.
Figures~\ref{fig:res_MNIST}, \ref{fig:res_KMNIST}, ~\ref{fig:res_SVHN} and~\ref{fig:res_Cifar10} report the accuracy of each of the compared methods after each annotation step on MNIST, KMNIST, SVHN and Cifar10 datasets respectively. See supplementary materials for more results.
{While most methods improve over random sampling, the margin of improvement is limited (2\%--3\%). When considering all the  studied datasets, {\tt Ours} and  {\tt Ours-app} performs similarly  to {\tt BALD},  {\tt MC-Dropout},   {\tt Err-Reductoin} and {\tt BADGE}.  It is worth noting that {\tt Ours-app}  is the  fastest to compute, as discussed in  supplementary.}
 \subsubsection{Imbalanced Setting.}   
\begin{figure*}[t!]
  \vspace{-0.2cm}
    \centering
    \subfloat[ \label{fig:res_Cifar10}]{{\includegraphics[width=.36\textwidth]{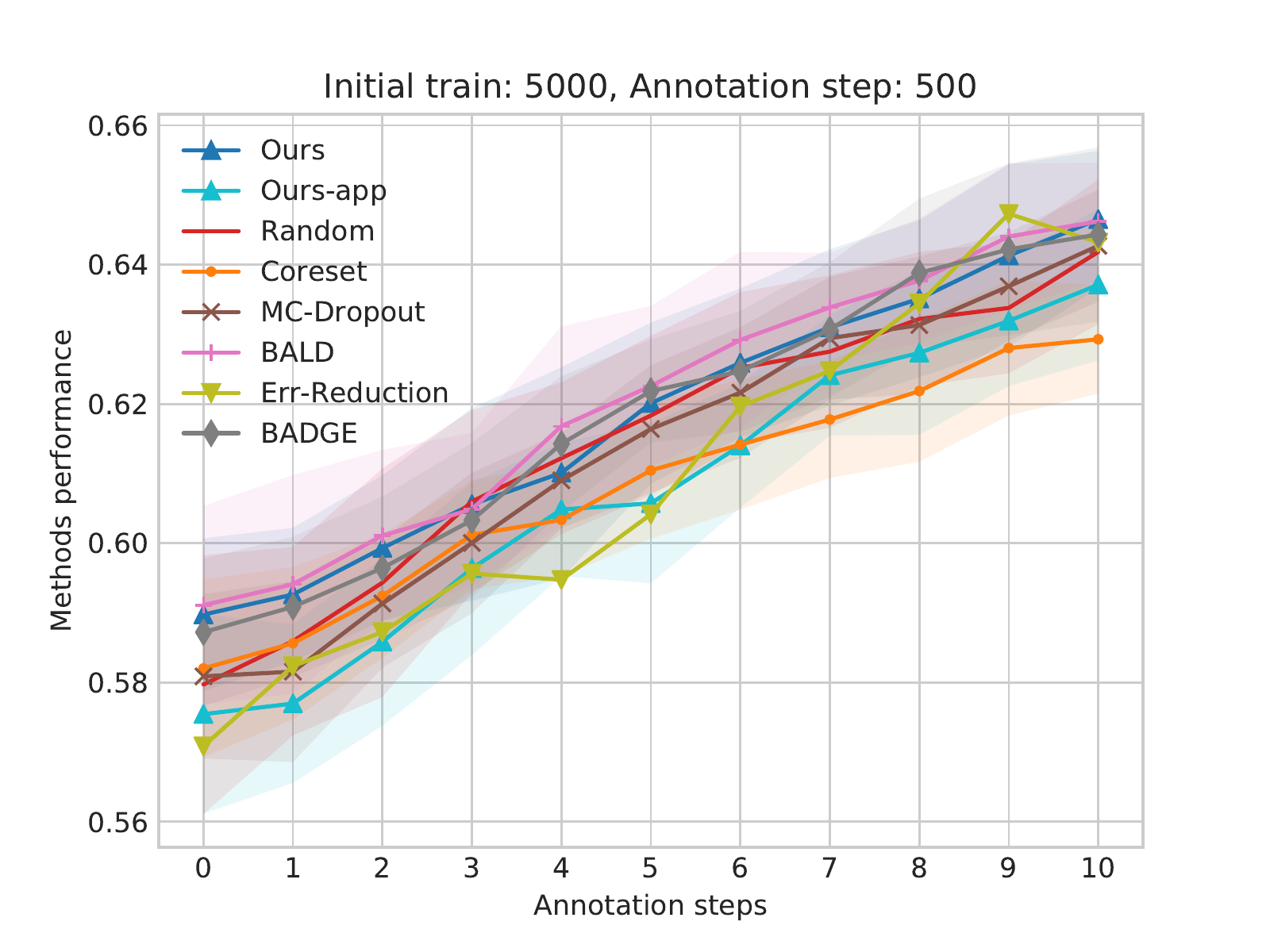} }}%
        \hfillx
    \subfloat[ \label{fig:res_imb_Cifar10}]{{\includegraphics[width=.36\textwidth]{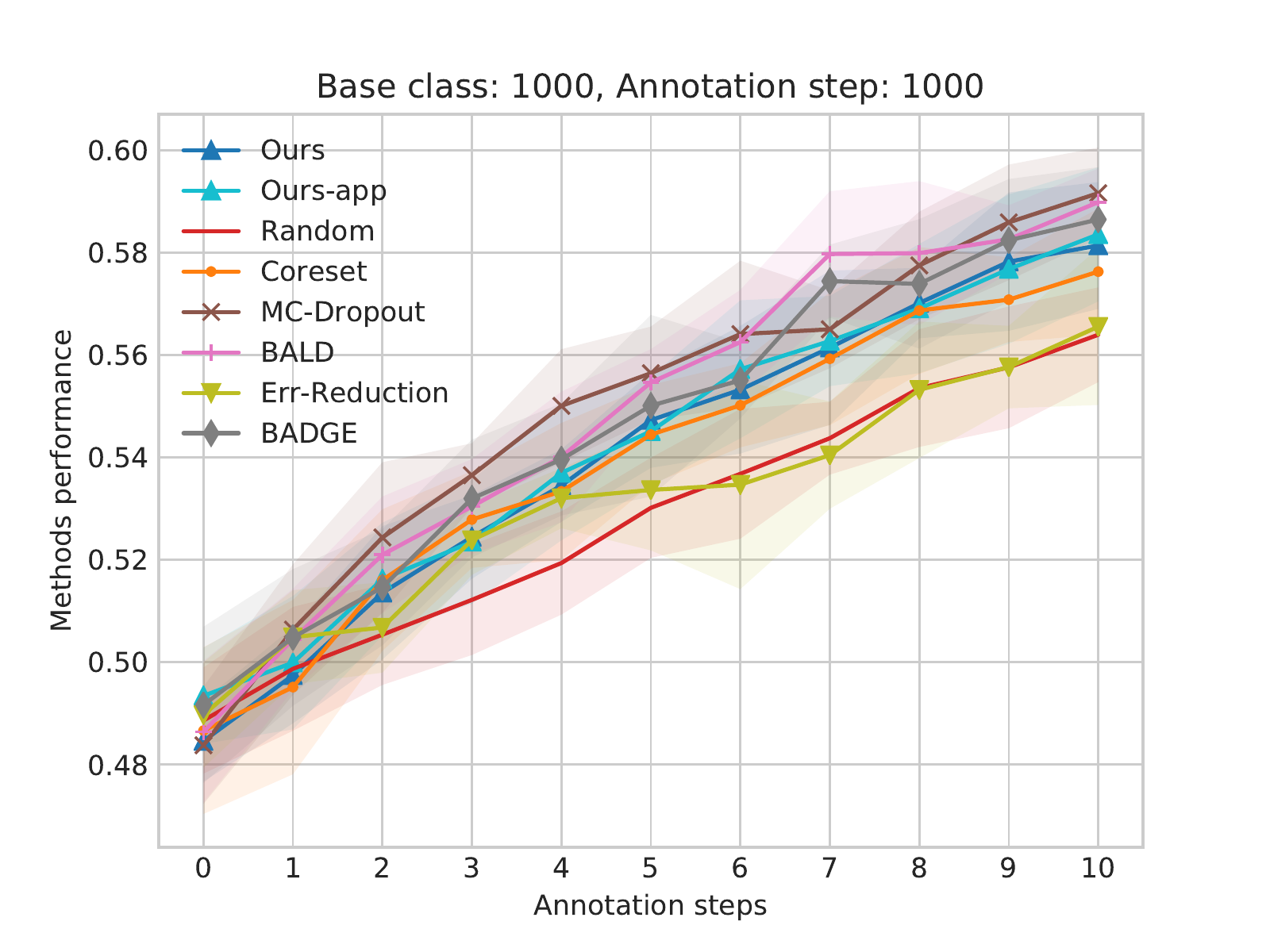} }}%
    \hfillx
  \subfloat[\label{fig:cityscapes_iou}]{{\includegraphics[width=0.33\textwidth]{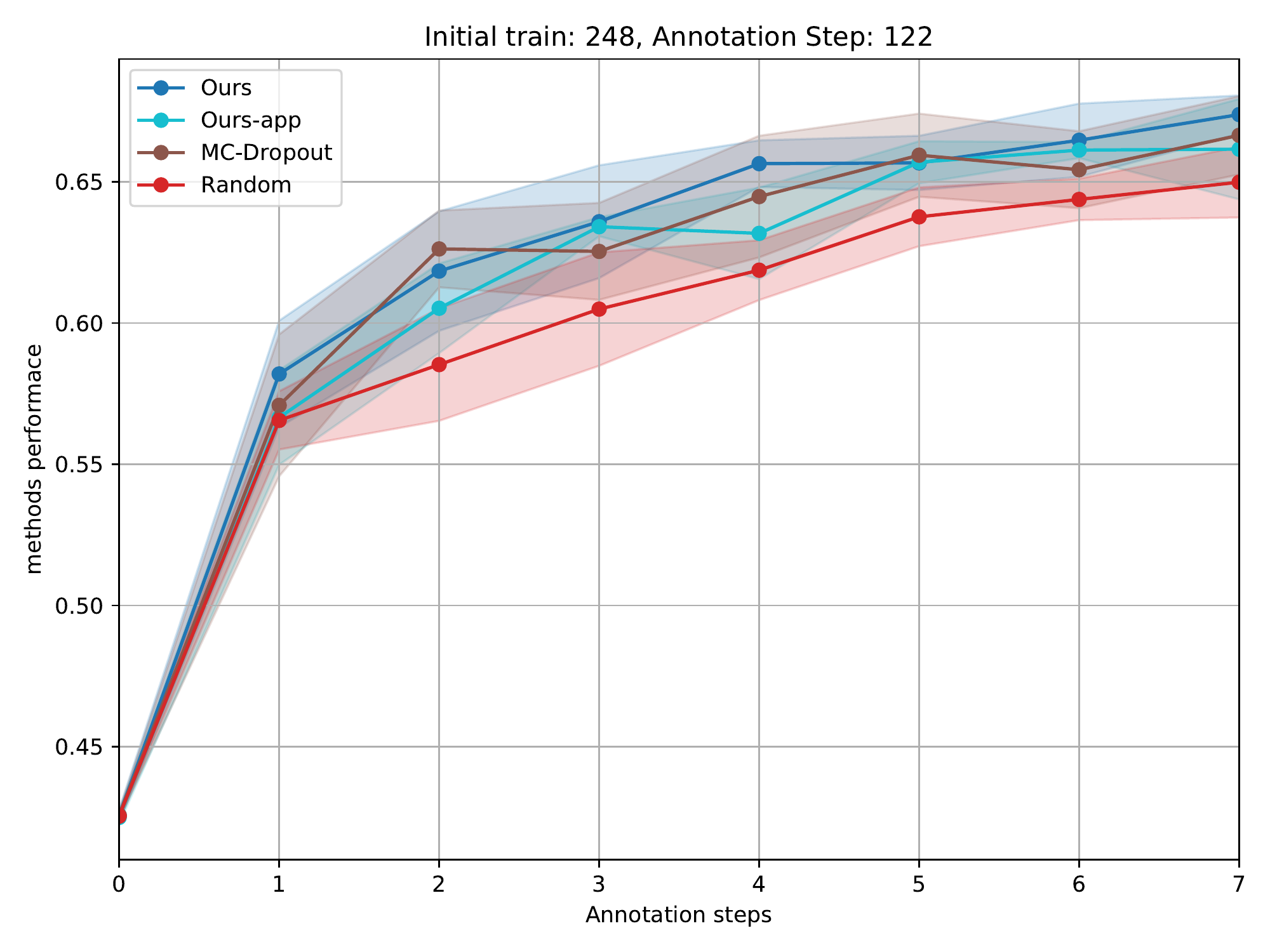}}}%
  \vspace*{-0.4cm}
   \caption{\footnotesize Mean accuracy and std.dev.  for (a)  Cifar10 balanced setting and (b)   Cifar10 imbalanced setting.   (c) Semantic segmentation results, mIoU and std.dev. on Cityscapes.}%
    \vspace{-0.4cm}
\end{figure*}

In the standard balanced setting, {\tt Random} appears to be a competitive baseline and only outperformed by a small margin as also shown in~\cite{sinha2019variational}.
Here, we argue that random sampling cannot be taken for granted as a competitive method in the cases where there are dominant categories that are not of much importance to the task at hand as opposed to the  under-represented ones.   For example, in autonomous driving applications most images contain examples of road, sky or buildings, but other categories like cyclists or trains are much less frequent. We simulate this scenario by constructing a pool of samples in which half of the categories are under-represented with number of samples equals to  $1/10$  of other categories samples (base  number of samples per class). Since the compared methods start with initial training set, we also construct this set with the same imbalanced setting. Regarding the validation set, we keep it balanced but limit its size to $1/5$ of the initial training size. We apply this setup to  the 4 studied datasets. As this is a much harder setting, for each dataset we double the base class size of the initial training set compared to the expected per category size in balanced setting, we also double the annotation step size.\\ 
Figures~\ref{fig:res_imb_MNIST},~\ref{fig:res_imb_KMNIST},~\ref{fig:res_imb_SVHN},~\ref{fig:res_imb_Cifar10} report the accuracy of each of the compared methods after each annotation step on the imbalanced sets of MNIST, KMNIST, SVHN and Cifar10  respectively, see supplementary materials for more results.
This setting  shows larger differences between the compared methods.  It is clear that {\tt Random} baseline fails to compete here, for example, on MNIST   {\tt Ours}  with only $5$  annotation steps achieves the same accuracy of {\tt Random} after $10$ annotation steps. This is a reduction of half of the annotation resources.  Uncertainty and information based methods {\tt BALD}, {\tt BADGE} and {\tt MC-Dropout} continue to improve over {\tt Random} with a larger  margin in this setting. {\tt Coreset}, in the contrary, has consistently lower performance, we think that this method is more suitable for extracting much larger batches of data. 
{\tt Ours}  achieves the best performance on MNIST, KMNIST and SVHN settings with a substantial margin.  We believe that this  is a significant improvement that shows our method ability to identify those underrepresented categories where most mistakes occur, and thereof selects most informative samples  contributing to higher gains in the trained model performance.
{ Notably, {\tt Ours}  is constantly outperforming  {\tt Err-Reduction}. This indicates that  using the largest generalization error as a proxy for identifying wrongly predicted samples provides the model with previously unknown knowledge and results in a larger reduction of the newly trained model generalization error  than when aiming at explicitly reducing it. }\\ 
For Cifar10 dataset, all methods perform closely; we think it is due to the low intra class similarity of this dataset  which reduces the potential impact of the individual selected samples. 
{\tt Ours-app}  improves over other methods on MNIST, KMNIST and  SVHN  benchmarks with slightly less margin than {\tt Ours}   and achieves comparable performance on Cifar10. This is also important results given its  low complexity.
\begin{table*}[h]
\vspace{-0.3cm}
  \begin{varwidth}[b]{0.35\linewidth}
    \centering
    \small
    \vspace{-1.5cm}
    \begin{tabular}{c c c c}
                 & Step 1 & Step 5 & Step 10  \\ \hline \hline
    {BALD}  & $72.2\%$ &$57.6\%$ & $56.6\%$  \\
    {MC-Dropout} &$70.4\%$  & $62.2\%$ & $60.6\%$  \\
    {Ours-app}  & $91.8\%$ & $67.2\%$ &$62.0\%$  \\
     {Ours}     & $90.0\%$ &$74.0\%$ & $70.0\%$ \\
    \end{tabular}
    \vspace{-0.4cm}
    \caption{ \footnotesize The percentage of wrong prediction among selected samples, on balanced MNIST   with 50 initial train and 50 annotation step. A larger annotation step  is used for a better estimation of  mistakes percentage.  }
    \label{tbl:mistakes}
  \end{varwidth}%
 \hfillx
  \begin{minipage}[b]{0.6\linewidth}
     \centering
  \subfloat[\label{fig:mnist_val_abl}]{{\includegraphics[width=0.5\textwidth]{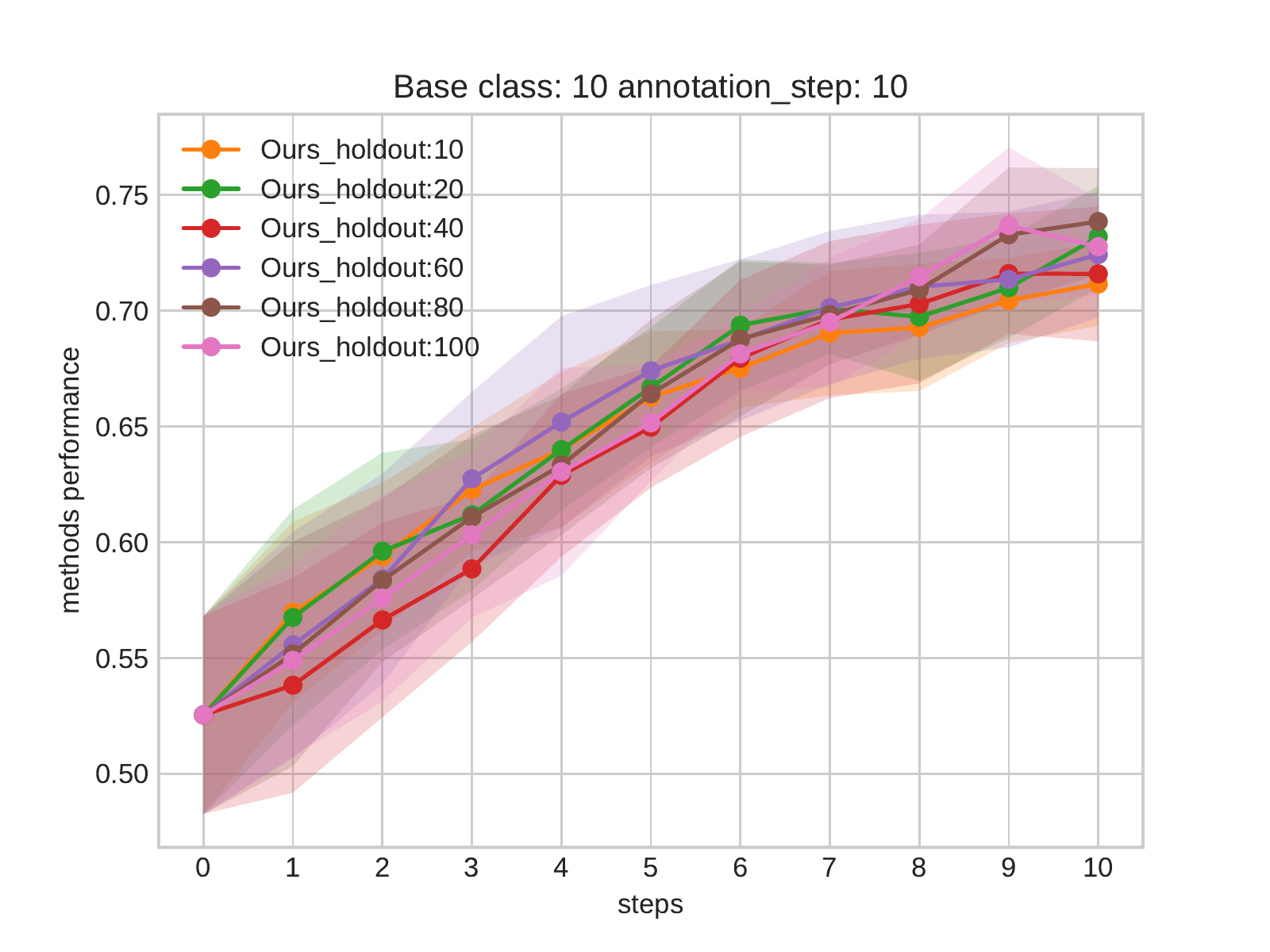} }}%
 \hfillx
\subfloat[\label{fig:svhn_val_abl}]{{\includegraphics[width=0.5\textwidth]{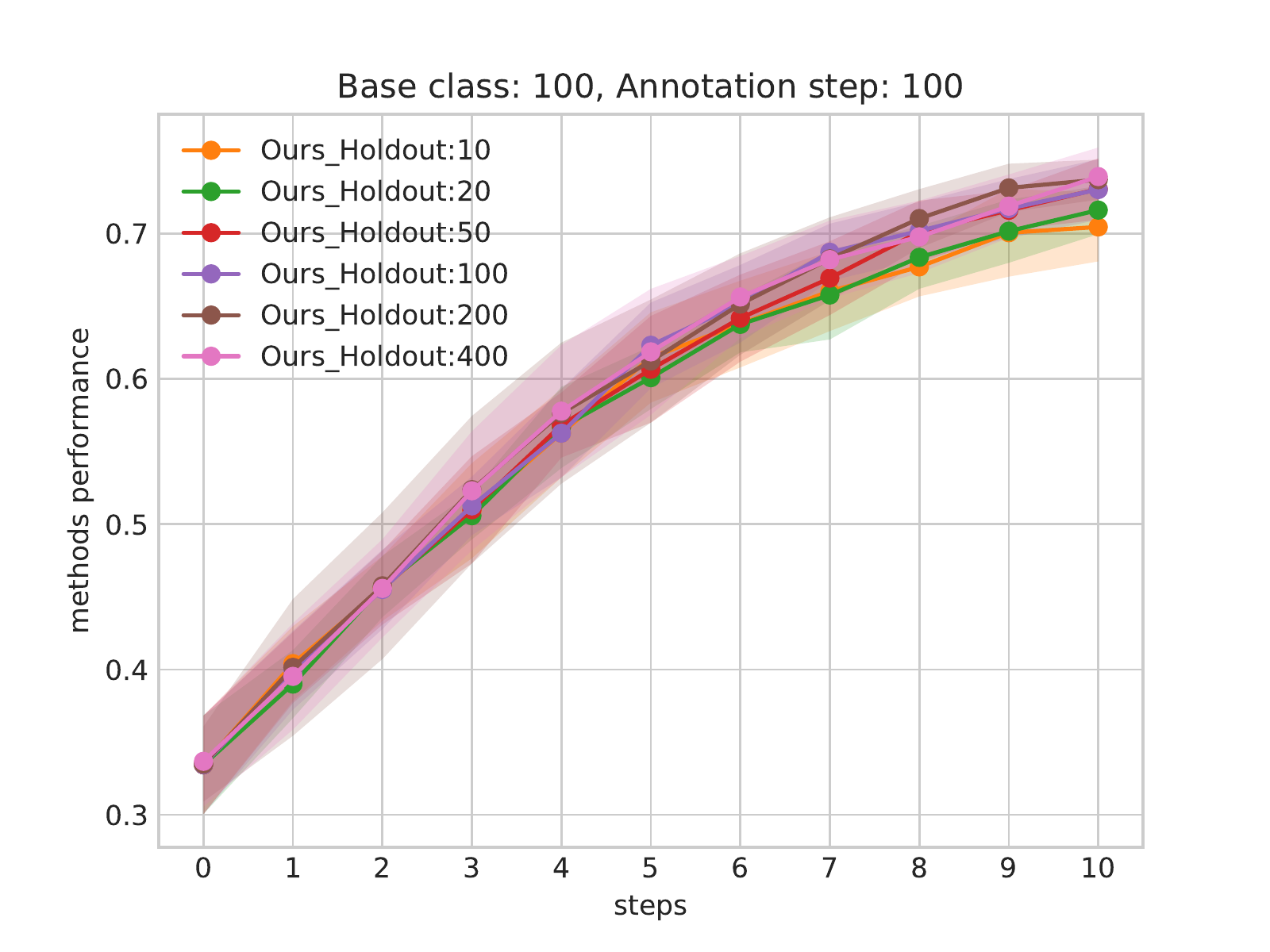} }}%
\vspace{-0.4cm}
   \captionof{figure}{ \footnotesize  Holdout set size ablation. \ref{fig:mnist_val_abl} is on Mnist imbalanced setting, \ref{fig:svhn_val_abl} is on SVHN imbalanced setting.  }
    \vspace{-0.35cm}
  \end{minipage}
\end{table*}
\vspace{-0.2cm}
\subsubsection{Holdout Set Size.}
{
Our criterion is based on the generalization error of the model $f(x;\theta)$, for $\theta=\theta_i^p$ -- the updated parameters of each pool sample, 
and depends on the used holdout set, regarded as representative of the different concepts (categories).  
 }
In all experiments, we have used the validation set for estimating our selection criterion. For  the imbalanced setting, the validation set was uniformly sampled but significantly smaller  than the size of the initial training  set. 
In this paragraph, we study empirically the effect of the holdout set size on the behaviour of our method.  
 Given an MLP backbone, Figure~\ref{fig:mnist_val_abl}  shows the performance of {\tt Ours}  using different holdout set sizes on imbalanced  MNIST, see supplementary  for balanced setting, while  Figure~\ref{fig:svhn_val_abl} reports the performance based on ResNet architecture and imbalanced SVHN. 
{ As it can be seen, only the very smallest holdout set sizes decay our method  performance. However, there doesn't seem to be a significant differences for larger sizes.
This empirical evidence  suggests that with only few samples $(>2)$ per class, our method can reach its best performance and in spite of the importance rule of this holdout set, its size doesn't affect significantly the performance of our method.  While one would expect that the more complex the ground-truth concept is, e.g. more complex images, the more holdout data is needed to estimate the generalization error; we argue that this is also the case of the typically deployed validation set.  We indeed require similar and even smaller size to rank the pool samples than what needed to tune the parameters of a neural network architecture. 
On another note, using training examples instead of holdout data, would affect negatively our criterion. Consider the case when  $\ell_v(\theta^p_i)$ is $0$ for each pool sample $x_i^p$, which could occur  when the model overfits its training data. In this case, the selection will be random.
In fact, in our approximation criterion~\eqref{eq:approximated_criterion}, we require a representative (non-zero) gradient~$\nabla_\theta\ell_v(\theta^s)$.
However,  with few holdout points the estimated  gradient direction  can be noisy but sufficient for the purpose of   candidates ranking.}
\vspace*{-0.2cm}
\subsubsection{Mistakes Selection.}
Our approach aims at  better identifying the wrongly predicted pool samples, which, once annotated, should improve the model performance, as it is being trained on previously unknown cases. 
We inspect the percentage of samples with wrong predictions in comparison to {\tt BALD} and {\tt MC-Dropout} as their criteria are related to the purpose of {\tt Ours}.
Table~\ref{tbl:mistakes} reports the accuracy in mistake selection  at the indicated steps in MNIST balanced setting.
Both our criteria are best at picking wrongly predicted samples, with {\tt Ours} achieving a higher mistake selection rate, $10\%$--$20\%$, than others. This behaviour was consistent on different settings/datasets. It further shows that our method outperforms the typical use of uncertainty and identifies better samples that are likely to be wrongly predicted.
\subsection{Semantic Segmentation Experiments}\label{sec:image_segmentation}
In the previous experiments, each sample has only one possible true class. We are interested in the case were each sample contributes to multiple and possibly conflicting hypotheses,
thus we consider semantic segmentation.\\
We mainly compare to {\tt Random} 
 and {\tt MC-Dropout} described previously,  Section~\ref{sec:image_classification}. 
For {\tt Ours} and {\tt Ours-app}: we only optimize the parameters or compute the gradients on the last two convolutional layers. We adapt {\tt Ours} to the case where an imperfect model produces both correct and incorrect predictions for one sample. We average pixel predictions $\mathcal{S}$~\eqref{eq:pseudo}, where $\ell\left(f(X^v;\theta^p_i),Y^v\right) > \ell\left(f(X^v;\theta^{s}),Y^v\right)$ holds, and by doing so we only consider the subset of the frame that indicates that the model has been negatively impacted by the pool sample. See supplementary  for further  details and results. 
Fig.~\ref{fig:cityscapes_iou} shows the mean Intersection over Union (mIoU) after each annotation step. {\tt Ours} and  {\tt MC-Dropout} performs closely, with {\tt Ours} having slightly higher  mIoU scores towards the end.
\section{Conclusion}\label{sec:conclusion}
We propose a new solution to the problem of active learning that first accepts  the hypothesis of the current model prediction on each pool sample and then judges the effect of increasing  this hypothesis confidence on the performance on a holdout set.  We use the change in the model generalization error as an indication of how likely the prediction of a given sample is to be mistaken. We further develop an approximation of our selection criterion and show that it  targets sample/prediction pairs that are dissimilar to those  in the holdout set from the current model perspective. We evaluate our approach on several benchmarks  and achieve comparable performance to  state-of-the-art methods. Additionally, { we setup for the first time a systematic compassion on the important and realistic imbalanced setting  where we show significant improvements}. Our method is computationally efficient and requires no changes on  the available model.
\section*{Ethical Impact}\label{sec:impact}
In general terms, active learning methods serve as enablers of machine learning models, and thus inherit similar ethical implications, perhaps accentuated by faster learning cycles. Other than the purpose and the ultimate application of the machine learning model, active learning can potentially have an impact on the use of data, and on human annotators. Regarding the latter, active learning aims at optimizing knowledge transfer from human experts to machine learning models, by directing attention to potentially meaningful samples. The implication would be that fewer resources are needed to achieve a task that sometimes may be considered as tedious and repetitive, but also that knowledge from those experts may eventually become irrelevant, as their expertise is superseeded by that of the collectively annotated data. As for the use of data, active learning methods, such as the one proposed in this work, are relevant to alleviate biases introduced in the data pool by an imperfect data collection process, or by the unbalanced nature of the true distribution of the data.


\bibliography{references/references.bib}
\end{document}